\pdfoutput=1

\documentclass[11pt]{article}

\usepackage[]{acl}

\usepackage{times}
\usepackage{latexsym}

\usepackage[T1]{fontenc}

\usepackage[utf8]{inputenc}

\usepackage{microtype}

\usepackage{graphicx}
\usepackage{times}
\usepackage{latexsym}
\usepackage{soul}
\usepackage{url}
\usepackage{amsthm}
\usepackage{booktabs}
\usepackage{bm}
\usepackage{amsmath}
\usepackage{tablefootnote}

\usepackage{multirow}
\usepackage[ruled,linesnumbered]{algorithm2e}
\usepackage{longtable}
\usepackage{array}
\usepackage{subfigure}
\usepackage{stfloats}
\usepackage{multicol}
\usepackage{color}
\usepackage{booktabs} 
\usepackage{arydshln}
\usepackage{balance}
\usepackage{xspace}
\usepackage{enumitem}
\usepackage{amssymb}

\newcommand{\hide}[1]{} 

\newcommand{\vpara}[1]{\vspace{1.5ex}\noindent\textbf{#1}}

\newcommand{\beq}[1]{\vspace{-0.03in}\begin{equation}#1\end{equation}}
\newcommand{\beqn}[1]{\begin{eqnarray}#1\end{eqnarray}}
\newcommand{\model}{NumericalTransformer}
\newcommand{\smodel}{NumericalTransformer }

\newcommand{\wmodel}{NT-NSM}
\newcommand{\swmodel}{NT-NSM }

\title{Injecting Numerical Reasoning Skills into Knowledge Base \\ Question Answering Models}

\author{Yu Feng$^{1,2}$, Jing Zhang$^{1,2}$\thanks{\text{ }\text{ }      Corresponding author.}, Xiaokang Zhang$^{1,2}$ , \\
\textbf{Lemao Liu$^{4}$, Cuiping Li$^{1,2}$, Hong Chen$^{1,2}$}\\
$^1$Key Laboratory of Data Engineering and Knowledge Engineering of Ministry of Education \\
$^2$School of Information, Renmin University of China\\
$^4$Tencent AI Lab
\\
\texttt{\{anniefeng,zhang-jing,zhang2718,licuiping,chong\}@ruc.edu.cn}\\
\texttt{\{lemaoliu\}@gmail.com}
}

\begin{document}

\maketitle

\begin{abstract}
Embedding-based methods are popular for Knowledge Base Question Answering (KBQA), but few current models have numerical reasoning skills and thus struggle to answer ordinal constrained questions. This paper proposes a new embedding-based KBQA framework which particularly takes numerical reasoning into account. We present \smodel on top of NSM, a state-of-the-art embedding-based KBQA model, to create \wmodel. To enable better training, we propose two pre-training tasks with explicit numerical-oriented loss functions on two generated training datasets and a template-based data augmentation method for enriching ordinal constrained QA dataset.
Extensive experiments on KBQA benchmarks demonstrate that with the help of our training algorithm, \swmodel is empowered with numerical reasoning skills and substantially outperforms the baselines in answering ordinal constrained questions. 
\end{abstract}

\section{Introduction}
\label{sec:intro}

Knowledge Base Question Answering (KBQA)~\cite{ZHANG202114} is to find the entities from the Knowledge Bases (KBs) such as Freebase~\cite{bollacker2008freebase} and DBPedia~\cite{lehmann2015dbpedia} as the answers of the factoid questions.
The knowledge in KBs is organized structurally as triplets in the $<$entity, relation, entity$>$ form, which can explicitly promote the answer reasoning process. Although one-hop questions can be solved with high accuracy~\cite{He_2021}, complex KBQA tasks which involve multi-hop reasoning, constrained reasoning, or numerical reasoning~\cite{bao-etal-2016-constraint,lan-jiang-2020-query,sun2020sparqa,kapanipathi-etal-2021-leveraging} are still challenging to be addressed.

Existing methods for both the simple and complex KBQA tasks can be generally categorized into semantic parsing-based (SP-based) methods~\cite{lan-jiang-2020-query,sun2020sparqa} and embedding-based methods~\cite{sun-etal-2018-open,saxena-etal-2020-improving,He_2021}. 
The SP-based methods translate a question into an executable
logic form in KBs, which is able to address various types of questions. However, these methods require the expensively annotated logic forms as supervision. Thus they are subject to narrow domains with a limited number of pre-defined logical predicates. 
Instead of parsing questions, the embedding-based methods represent the questions and entities in the KBs together and then select the entities that are most relevant to the question. These methods can be trained end-to-end without the supervision of the logic forms. As a result, they are more fault-tolerant to different domains.
Despite the advantages of such embedding-based methods, most of them focus on solving the single- or multi-hop questions. To answer the complex question ``What is Taylor Swift’s latest album?'' in Figure~\ref{fig:framework}, 
the answer ``Evermore'' is supposed to encode not only the magnitude of its album release date but also the ordinal relationship with ``latest''---the ordinal determiner in the question. Existing embedding-based models are not explicitly aware of the magnitude and ordinal properties of entities, making the entity representations fall short in the ability to support such numerical reasoning.

In view of the above issue, this paper targets empowering the embedding-based KBQA models with the skills of numerical reasoning to address the ordinal constrained questions. 
The ordinal constraint is summarized as one of the most important constraints via web query analysis~\cite{bao-etal-2016-constraint} and ordinal is also defined as the second fundamental measurement to capture data in the forms of surveys\footnote{https://www.questionpro.com/blog/nominal-ordinal-interval-ratio/}. 
Existing embedding-based models lack such numerical reasoning skills because their created number embeddings can not encode two critical properties:

\begin{itemize}
    \item \textbf{Relative Magnitude} between numbers, such as ``$1\prec2\prec3$'' \footnote{We use ``$1\prec2\prec3$'' to denote that the magnitude between 1 and 2 is closer than that between 1 and 3 instead of ``$1<2<3$'' because the embeddings can only reflect the relative distance rather than the absolute magnitude.}.
    \item \textbf{Ordinal relationship} between each number and the ordinal determiner. For example, 3 in ``$1\prec2\prec3$'' is identified as the largest number for the ordinal determiner ``largest''. 
\end{itemize}

Number embeddings which satisfy the above two properties are capable of numerical reasoning for ordinal constrained questions. To obtain such number embeddings, in our previously work~\cite{feng-etal-2021-pretraining-numerical}\footnote{This paper is the extension of the previous work published in EMNLP2021~\cite{feng-etal-2021-pretraining-numerical}.}, we propose two modules together with two separated self-supervised loss functions for pre-training them. Specifically, the former module NumGNN is a GNN model built upon a constructed number graph with nodes as numbers and edges pointing from the larger numbers to the smaller numbers. Such a number graph that distinguishes the magnitude of
numbers together with a number-aware triplet loss can make the GNN model encode the relative magnitude in its output number embeddings. The latter module NumTransformer is a vanilla transformer-structured model whose input is the question embedding and the number embeddings outputted by the above GNN model. Through optimizing a number prediction loss function, this transformer can encode the ordinal relationship between the numbers and the ordinal determiners in its output number embeddings. 

The final number embeddings outputted by the latter transformer-structured model can be injected into the entity embeddings learned by any embedding-based  KBQA  model with the devised attention mechanism. As a result, the fused entity embeddings can be empowered with numeric reasoning skills.
The experiments on two benchmarks of KBQA, i.e.,  WebQSP and CWQ, show that NumGNN plus NumTransformer, serving as plugins of alternative embedding-based KBQA models, can achieve substantial and consistent improvement (+2.4 -14.8\% in terms of accuracy) on the ordinal constrained questions compared with the same models removing the plugins.

Based on the previously published models and results, this paper makes further improvements to overcome two main drawbacks. First, encoding number embeddings depends on two models, which demands much training as well as inference efforts. Second, because of the unrigorous fusion operation, the existing method may inject number embeddings for questions of non-numerical-reasoning type, or fusing number embeddings into irrelevant entities to the question, which will lead to worse answers.

In view of the above limitations of our previous work, this paper reduces the two models into a single model named \underline{N}umerical\underline{T}ransformer (NT), which is a transformer-structured model being injected with a specially devised self-attention mask matrix serving as ``NumGNN'' to preserve the relative magnitude.
By optimizing the same number-aware triplet loss as well as the number prediction loss, this single NumericalTransformer model can be pre-trained to preserve both the relative magnitude and the ordinal relationship, achieving even better performance than the previous proposed two models. 

After pre-training the NumericalTransformer, to better inject the outputted number embeddings into the entity embeddings learned by any embedding-based KBQA model, we propose an enhanced comprehensive reasoning module, which is empowered with 
a question type classifier instead of the previous rule-based method to identify the questions requiring the number injection, as well as 
an additional entity pruning strategy to locate the most relevant entities and numbers to the question for injecting the number embeddings. 

Based on the enhanced framework, we propose \swmodel which is instantiated by NSM~\cite{He_2021}, a state-of-the-art embedding-based KBQA model. 
\swmodel can be learned by the end-to-end QA supervision signals.
Since such QA supervision signals are rarely available in existing KBQA datasets, a template-based data augmentation approach is also proposed to enrich the ordinal constrained QA pairs for sufficient training.

\begin{enumerate}
\item \swmodel is able to address the task of ordinal constrained KBQA well.
\item The framework of \swmodel can be adapted to other embedding-based KBQA models.
\item Pre-training tasks on the NumericalTransformer provide the number embeddings with numerical reasoning properties.
\item The augmented pseudo ordinal constrained QA data improves the QA performance.
\end{enumerate}

In summary, the main contributions of this work include: 

\begin{itemize}
\item An embedding-based KBQA framework for particularly considering numerical reasoning.
\item A pre-training strategy with the generated training data and specifically designed loss functions for explicitly supervising number embedding learning.
\item A template-based data augmentation strategy for generating ordinal constrained KBQA examples as supervision.
\end{itemize}

\section{Related Work}
\vpara{Knowledge Base Question Answering.}
SP-based and embedding-based methods are two main kinds of methods developed for KBQA~\cite{HeIJCAI21,ZHANG202114}. SP-based methods~\cite{berant-etal-2013-semantic,berant-liang-2014-semantic,yih-etal-2015-semantic,bao-etal-2016-constraint,liang-etal-2017-neural,lan-jiang-2020-query,sun2020sparqa} target at parsing the questions into the executable logic forms in the KBs. They can address complex questions, but suffer from a small number of logic predicates or require expensively annotated logic forms as supervision. 
Instead of parsing the question, embedding-based methods~\cite{miller-etal-2016-key, sun-etal-2018-open,saxena-etal-2020-improving, He_2021} embed the entities in the whole KB~\cite{miller-etal-2016-key,saxena-etal-2020-improving} or only embed the entities in the extracted question-relevant subgraph~\cite{chen2019bidirectional,He_2021,sun-etal-2018-open}, and then rank the entities according to their relevance to the question. 
They are more fault-tolerant due to the robustness of neural models but the recent methods are incapable of addressing the ordinal constrained questions.

\vpara{Numerical Reasoning.}
Numerical reasoning has been studied for Machine Reading Comprehension (MRC)~\cite{yu2018qanet, wallace2019nlp, zhang2020language,geva2020injecting}. 
Wallace et al.~\cite{wallace2019nlp} find BERT~\cite{devlin2019bert} has limited capacity in numerical reasoning. 
NumBERT~\cite{zhang2020language} measures numeracy by scientific notations and designs the specific scalar probing task to fine-tune BERT. GenBERT~\cite{geva2020injecting} further modifies BERT's structure for this purpose.
Injecting the numerical reasoning skills in KBQA are different from MRC, because: (1) MRC can easily develop a unified encoder for the textual question and passages. However, KBQA takes more efforts on bridging the gap between the textual question and structured KB, which increases the numerical reasoning difficulty. (2)
The numbers that impact the answer are exactly included in the given passage of MRC. However, for KBQA, numbers relevant to the question must be carefully located to keep the noisy numbers out.

\section{Problem Definition}
\label{sec:problem}

\noindent A \textbf{Knowledge Base (KB)} $\mathcal{G}$  organizes factual information as a set of triples,  $\mathcal{G}=\{(e,r,e')|e,e'\in \mathcal{E}, r\in \mathcal{R}\}$, where $\mathcal{E}$ and $\mathcal{R}$ denote the entity set and the relation set respectively. 
A relation $r$ falls into two types: a numerical relation if any of its tail entities is a number and a non-numerical relation otherwise. The tail entity of a numerical relation is named as a numerical value and denoted by $v$.
Let $\textbf{E}\in \mathbb{R}^{|\mathcal{E}| \times d }$ and $\textbf{R}\in \mathbb{R}^{|\mathcal{R}| \times d}$ denote the embedding matrices for entities and relations respectively where each row vector $\textbf{e} \in \mathbb{R}^d$ or $\textbf{r} \in \mathbb{R}^d$ is the embedding for entity $e$ or relation $r$ initialized by RoBERTa~\citep{liu2019roberta}. 

\vpara{Knowledge Base Question Answering (KBQA):} 
Given a question $q$ and a KB $\mathcal{G}$, KBQA targets at extracting the answer entities $a_q$ from $\mathcal{E}$. The topic entities are entities mentioned in the question. 

\vpara{Ordinal constrained KBQA}~\citep{bao-etal-2016-constraint}: An ordinal constrained question denotes that the answers of such question should be selected from a ranked set based on the ordinal determiner in the question as ranking criteria. 
The typical ordinal determiner can be biggest, latest, etc. KBQA for such questions is ordinal constrained KBQA. 


 \begin{figure*}[t]
	\centering
	\includegraphics[width=0.9\textwidth]{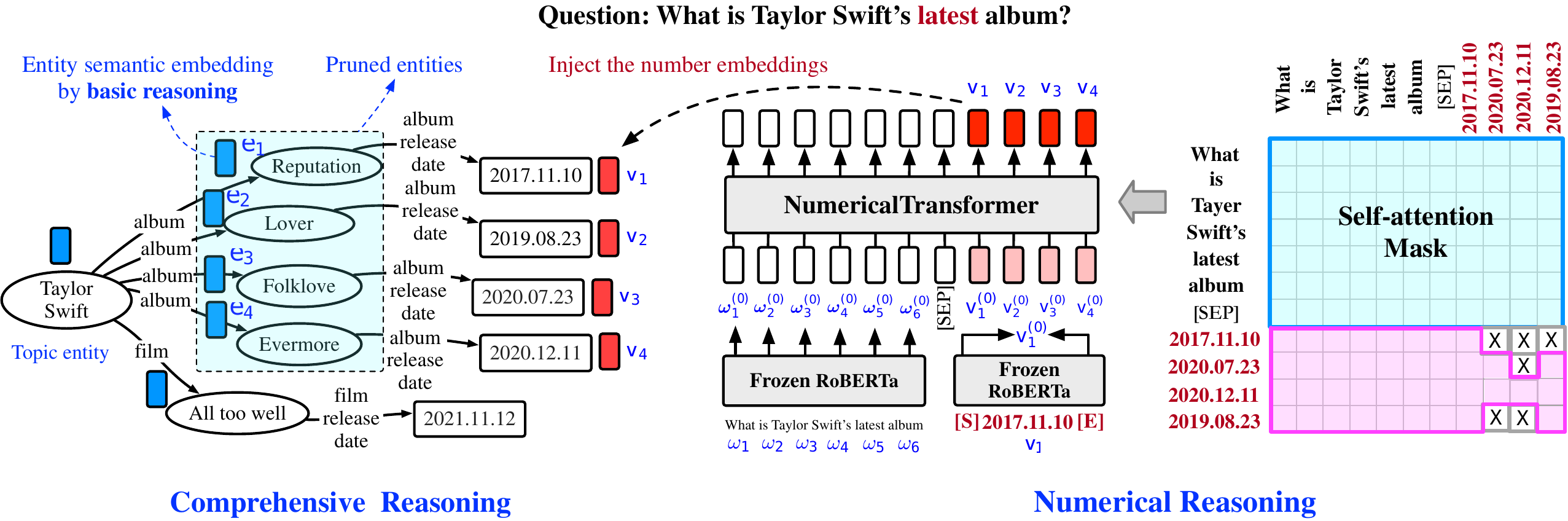}
	\caption{\label{fig:framework} We perform numerical reasoning by \smodel on the question and the related numbers, and inject the number embeddings into the entity embeddings inferred by the basic reasoning. A built-in self-attention mask controls the attention mechanism, where a number can attend to the question words and the smaller numbers.}
\end{figure*}

\section{ Embedding-based KBQA Framework}
\label{sec:model}

\vpara{Overall Framework} is presented in Figure~\ref{fig:framework}, which consists of basic reasoning instantiated by NSM~\cite{He_2021}, numerical reasoning and comprehensive reasoning modules. Specifically, NSM first takes the responsibility of \textbf{basic reasoning}, which learns entity embeddings to encode the semantic relationships with the textual question. For example, to answer the question ``What is Taylor Swift’s latest album?'', the embeddings for albums should be closely related to the question embedding.
Then \smodel targets at \textbf{numerical reasoning}, which learns the number embeddings for the numerical values of the entities. For example, it needs to infer the embeddings for the release dates of albums to encode relative magnitude between dates and the ordinal relationship with ``latest''.
Finally, a \textbf{comprehensive reasoning} module is required to inject the number embeddings into entity embeddings to complete ordinal constrained KBQA. 

\subsection{Basic Reasoning}
\label{sec:basicreasoning}
We adopt NSM~\cite{He_2021}, a state-of-the-art embedding-based KBQA model for basic reasoning.
The main idea is to perform a GNN-based algorithm on the two-hop topic-entity-centric subgraph for extracting the answers.
We formulate the process as:
\beq{
 \label{eq:basicembedding}
 \textbf{E} =  {\rm NSM}( \mathcal{G}_q, \textbf{E}^{(0)},\textbf{R}^{(0)}; \Phi),
}

\noindent where $\mathcal{G}_q$ is the subgraph, $\mathcal{E}_q$ are the entities in $\mathcal{G}_q$. We only include the non-numerical relations between entities into $\mathcal{G}_q$. $\textbf{E}^{(0)}$ and $\textbf{R}^{(0)}$ are the initial embeddings of the entities and relations. $\Phi$ denotes the parameters of NSM. The output entity embeddings can encode the semantics of the entities and are to predict the probability of $e_i$ being the answer:
\beq{
\label{eq:basicpredict}
p(e_i \vert q,\mathcal{G}_q)={\rm \sigma}(\textbf{W}_{\rm  predict}^\top \textbf{e}_i + b_{\rm predict}),
}

\noindent where $\textbf{W}_{\rm  predict}$ and $b_{\rm predict}$ are parameters for prediction.
For stable training of \wmodel, we pre-train NSM as stated in the origin paper.

\subsection{Numerical Reasoning}
\label{sec:numericalreasoning}
For numerical reasoning, we propose \smodel which accepts the question and the numerical values as input. 
To deal with any number, we treat a numerical value as a sequence of tokens and apply RoBERTa to initialize it.
To capture the relative magnitude, we devise a self-attention mask matrix in the transformer encoder to constrain the interactions between the numbers.
Below we explain (1) \textbf{the input} and (2) \textbf{the encoder}.

\vpara{(1) The Input with Special Number Embeddings (SNE).}
The input of \smodel is comprised of the question and the numbers (\emph{i.e.}, numerical values) belonging to the same numerical relation that is relevant to the given question. 
Specifically, we retrieve the top-$K$ relevant numerical relations $\mathcal{R}_q$ via measuring the cosine similarity between the  question embedding and relation embedding. 
A question $q=\{\omega_1, \omega_2, \cdots, \omega_n\}$ is a sequence of words. We encode it by RoBERTa, output the initial embedding $\bm{\omega}_i^{(0)}$ for each word $\omega_i^{(0)}$ and average them as the question embedding $\textbf{q}$.
For each relation $r \in \mathcal{R}_q$, we extract the corresponding numerical values of the entities in $\mathcal{G}_q$ to create the number set $\mathcal{V}_r = \{v | (e,r,v)\in\mathcal{G}, e \in \mathcal{E}_q\}$.
$\mathcal{V}_r$ of different $r$ composes $\mathcal{V}_q$. 

We propose a special number embedding method to initialize each number $v_i \in \mathcal{V}_r$. Specifically, we treat each $v_i$ as a sequence of tokens and enclose it with the same start token [S] and end token [E] to create ``$\mbox{[S]}v_i\mbox{[E]}$''. We use RoBERTa to encode it and concatenate the output embeddings of [S] and [E] as $v_i$'s initial embedding $\textbf{v}_i^{(0)}$. 
Note in this paper, we use the original published parameters of RoBERTa, which is frozen in our model.
Formally, the input $\textbf{H}^{(0)}$ is the concatenation of the initial embeddings of the question words and all the numbers in a number set $\mathcal{V}_r$, \emph{i.e.}, 
\beqn{
\label{eq:transformerinput}
\textbf{H}^{(0)} \!\!=\! \bm{\omega}_1^{(0)};\cdots;\bm{\omega}_n^{(0)};\mbox{[SEP]};\textbf{v}_1^{(0)};\cdots;\textbf{v}_m^{(0)},
}

\noindent where $\bm{\omega}_i^{(0)}$ and $\textbf{v}_j^{(0)}$ are the initial embeddings of the $i$-th word and the $j$-th number respectively. Since a numerical value is treated as a token sequence, we can easily embed any unseen values online by RoBERTa instead of maintaining a number embedding for each number offline.

\vpara{(2) The Encoder with Self-Attention Mask Matrix (SAM).}
We adopt a transformer of $L$ layers to update the embeddings of the input tokens. Each layer consists of a multi-head self-attention layer and a fully connected feedforward network. A self-attention head in layer $l$ is defined as:
\begin{align}
\label{eq:attention}
&\textbf{Q}^{(l)}=\textbf{H}^{(l)} \textbf{W}_Q^{(l)}, \textbf{K}^{(l)}= \textbf{H}^{(l)}\textbf{W}_K^{(l)}, \textbf{V}^{(l)}= \textbf{H}^{(l)}\textbf{W}_V^{(l)},
\notag\\ 
&\textbf{A}^{(l)}={\rm softmax} \left(\frac{\textbf{Q}^{(l)}(\textbf{K}^{(l)})^{\top}}{\sqrt{d_k}} +\textbf{M}   \right)  \textbf{V}^{(l)},
\end{align}

\noindent where $\textbf{W}_Q^{(l)},\textbf{W}_K^{(l)},\textbf{W}_V^{(l)} \in \mathbb{R}^{d_h \times  d_k}$ are model parameters. We use $\Theta$ to represent all these parameters. $\textbf{M} \in \mathbb{R}^{(n+m+1) \times (n+m+1)}$ is the matrix of self-attention mask.
For the number parts, we set $M_{ij} = 0$ to allow token $i$ to attend to token $j$ if the $i$-th token is larger than the $j$-th token, and set $M_{ij} = -\infty$ for prevention. 
Take Figure~\ref{fig:framework} as an example, ``2020.07.23'' can only interact with ``2017.11.10'' and ``2019.08.23''.
As a result, all the numbers can interact with the question words, but they can only attend to the smaller numbers. This mechanism implicitly distinguishes the small numbers from the large ones as the embeddings of the large ones encode more information via more interactions than the embeddings of the small ones.

\subsection{Comprehensive Reasoning}
\label{sec:comprehensivereasoning}
Comprehensive reasoning aims to inject the above learned number embeddings with the entity embeddings learned by basic reasoning. During this, we need to devise (1) how to integrate the number embeddings (\textbf{number embedding integration}) (2) on which entities (\textbf{entity pruning}) for the (3) ordinal constrained questions (\textbf{question classifier}). Note number embedding integration adopts the same technique of our previous work~\cite{feng-etal-2021-pretraining-numerical}.

\vpara{(1) Number Embedding Integration.} 
The semantic embedding $\textbf{e}_i$ for any $e_i$ in $\mathcal{G}_q$ is inferred by the basic reasoning module (Eq.~\ref{eq:basicembedding}). Its number embedding $\widetilde{\textbf{e}}_i$ can be obtained by aggregating the number embeddings of all the numerical values of the interest, \emph{i.e.},
\begin{align}
\label{eq:numberplugin}
& \widetilde{\textbf{e}}_i \;=\!\!\sum_{(r_j, v_j)\in \mathcal{N}_{\rm num}(i)} \!\! \alpha_j (\textbf{W}^{\top}_{\rm num}[\textbf{r}_j;\textbf{v}_j]+\textbf{b}_{\rm num}) , \nonumber\\ 
 & \alpha_j = {\rm softmax}( \textbf{r}_j^{\top} \textbf{q}), \end{align}

\noindent where $\mathcal{N}_{\rm num}(i) = \{(r_j, v_j) \vert(e_i, r_j, v_j) \!\in\! \mathcal{G},  r_j \!\in\! \mathcal{R}_q \}$ is $e_i$'s numerical relations and the corresponding values. $\textbf{r}_j$ output by RoBERTa and $\textbf{v}_j$ output by \smodel are the embeddings for $r_j$ and $v_j$ respectively. $\alpha_j$ measures the relevance between $r_j$ and $q$, which emphasizes the effect of the question-relevant numbers. 
Then $\textbf{e}_i$ and $\widetilde{\textbf{e}}_i$ are integrated together as:
\beqn{
\label{eq:finalembedding}
\hat{\textbf{e}}_i = \textbf{W}_{\rm  inter}^{\top}([\textbf{e}_i;\widetilde{\textbf{e}}_i]) + \textbf{b}_{\rm inter},
}

\noindent where $\widetilde{\textbf{e}}_i$ is set to $\textbf{0}$ if $e_i$ has none numerical values.
Then we estimate the probability of $e_i$ being the answer considering their numerical properties as:
\beqn{
\label{eq:comprehensivepredict}
p(e_i \vert q,\mathcal{G}_q, \mathcal{R}_{q}, \mathcal{V}_q)={\rm \sigma}(\textbf{W}_{\rm pred}^\top \hat{\textbf{e}}_i + b_{\rm pred}),
}

\noindent where $\textbf{W}_{\rm num}$, $\textbf{b}_{\rm num}$, $\textbf{W}_{\rm inter}$, $\textbf{b}_{\rm inter}$, $\textbf{W}_{\rm pred}$, and $b_{\rm pred}$ compose the parameters $\Psi$ of comprehensive reasoning.

\vpara{(2) Entity Pruning.} 
The above integration performed on any entities in $\mathcal{G}_q$ may inject noises. For example, in Figure~\ref{fig:framework}, although the entity ``All too well'' is a film rather than an album, it has a numerical relation ``film release date'' which is also relevant to the question. Following the above integration strategy, the final embedding of ``All too well'' is also fused with the number embedding of ``2021.11.12'', which may wrongly boost the predictive probability of ``All too well'' as the answer. To reduce the noisy integration, we only inject numbers into the entities that are highly relevant to the question. We resort to the pre-trained basic reasoning module for entity pruning. Specifically, we select entities whose $p(e_i \vert q,\mathcal{G}_q)$ estimated by Eq.~\ref{eq:basicpredict} are larger than $\mu$ as the pruned candidates $\mathcal{C}$, and only inject the number embeddings for $\mathcal{C}$. In this way, each $\mathcal{V}_r$ can be reduced into $\{v \vert (e_i, r, v) \in \mathcal{G}, e_i \in \mathcal{C}\}$, which can also improve the inference efficiency of \model. 

\vpara{(3) Question Classifier.}
We perform the whole reasoning process if a question is ordinal constrained but only basic reasoning otherwise. To enable different reasoning choices, we train a question type classifier by RoBERTa, which uses the [CLS] embedding of a question to predict the type probability. 
Based on it, we update the predictive probability of an entity $e_i$ being the answer as:
\beqn{
\label{eq:typeprediction}
p( e_i \vert q, \mathcal{G}_q)\!\!\! &=& \!\!\!p(t_q = \mbox{O}) p(e_i \vert q,\mathcal{G}_q,\mathcal{R}_q, \mathcal{V}_q) \\
\!\!\!&+& \!\!\!(1-p(t_q = \mbox{O})) p(e_i \vert q,\mathcal{G}_q), \nonumber
}

\noindent where $p(t_q  = \mbox{O})$ indicates the probability of the question type $t_q$ being ordinal constrained.

\vpara{Loss Function.}
We learn the parameters $\Phi$ for basic reasoning, $\Theta$ for numerical reasoning, and $\Psi$ for comprehensive reasoning by optimizing the cross-entropy loss between the predictive probabilities and the ground truth answers.

 \begin{figure*}[t]
	\centering
	\includegraphics[width=0.85\textwidth]{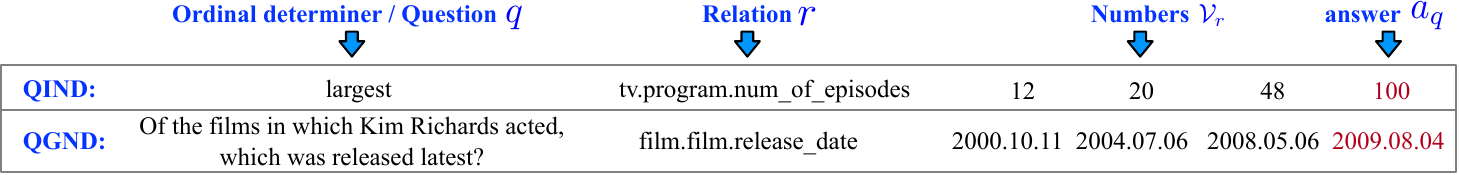}
	\caption{\label{fig:numericaldata} Two kinds of automatically generated datasets for pre-training \model.}
\end{figure*}

\section{Pre-training and Data Augmentation}
\label{sec:training}
\subsection{Pre-training Tasks}
\label{sec:pretraining}
Although the \smodel in \swmodel allows the rich interactions between the question words and the numbers, the number embeddings are learned implicitly without the explicit annotation of the two numerical properties in the standard KBQA training data. To enhance the number embeddings, we propose two pre-training tasks according to the two numerical properties by generating two kinds of training data $\{( q, r, \mathcal{V}_r, v_q)\}$
and defining two objective functions on top of the training data. The \smodel is only pre-trained once on the generated training data and then it can be adapted to all KBQA datasets and basic reasoning models.
The two objective functions are the same as our previous work~\cite{feng-etal-2021-pretraining-numerical}, but are used for the unified NumericalTransformer instead of the two modules of the previous work. Due to the unified model, a more comprehensive data construction method for 
generating richer datasets, QIND and QGND, is devised for pre-training the proposed NumericalTransformer.


\vpara{Question-irrelevant Numerical Dataset (QIND).}
Considering the difficulty of creating a textual question, we generate a question-irrelevant dataset on a large scale.
Specifically, we randomly extract a set of numerical relations from $\mathcal{G}$ and sample $N$  numbers associated with each relation $r$ into $\mathcal{V}_r$. Then we choose an ordinal determiner $q$ and annotate the number in $\mathcal{V}_r$ satisfying $q$ as the correct answer $v_q$. The tuple $(q,r,\mathcal{V}_r,v_q)$ composes a training instance. The ordinal determiners include ``largest'', ``fewest'', ``biggest'', and ``smallest'' for size-based numerical relations and ``earliest'', ``latest'', ``most recent'', ``first'', and ``last'' for time-based numerical relations. 
Figure~\ref{fig:numericaldata} illustrates a generated instance, where ``12 20 48 100'' are the numbers associated with the relation ``tv.program.num\_of\_episodes''. ``100'' can be automatically annotated as the correct answer guided by the ordinal determiner ``largest''. 

\vpara{Question-guided Numerical Dataset (QGND).}
We create a second question-guided dataset to enhance the connection with authentic data. Specifically, we retrieve ordinal constrained QA pairs from CWQ's training/validating dataset.\footnote{Other datasets indicating the relationship between ordinal determiners and numbers can also be chosen for creating QGND.}
For each QA pair $(q,a_q)$, we select the relevant $r$ that can derive the answer, and include the number $v$ satisfying $(a_q, r, v) \in \mathcal{G}$ as the correct number $v_q$ into $\mathcal{V}_r$. Then we add several other numbers associated with $r$ into $\mathcal{V}_r$. Figure~\ref{fig:numericaldata} illustrates an instance.


\vpara{Pre-training Objectives.}
We propose a number-aware triplet loss (NTL) and a number prediction loss (NPL) and optimize them on QIND first and then on QGND.
NTL aims to preserve the relative magnitude of numbers. For each instance $( q, r, \mathcal{V}_r, v_q )$, we randomly sample three numbers from $\mathcal{V}_r$ and allow the small number $v_s$ to be closer to the median one $v_m$ than the big one $v_b$. In another word, ``$v_s \prec v_m\prec v_b$''  should be satisfied.
The loss function is:
\beq{
\small
\label{eq:numbertripletloss}
\ell^1 =\sum_{(v_s,v_m,v_b) \in \mathcal{T}}{\rm max}(0,\epsilon+ g(\textbf{v}_s,\textbf{v}_m)-g(\textbf{v}_s,\textbf{v}_b)),
}
\normalsize

\noindent where $g$ is the cosine similarity between two numbers, $\textbf{v}$ is the output number embedding of \model, $\mathcal{T}$ is the set of the sampled triplets, and $\epsilon$ is a margin separating $(v_s,v_m)$ and $(v_s,v_b)$.

NPL captures the ordinal relationship.
For example, we aim to make the embedding of 1 in ``$1\prec2\prec3$'' 
 closer to the ordinal determiner ``smallest'' than 2 and 3. 
For a generated instance, we predict the probability of $v_i$ being the answer as:
\beq{
\small
\label{eq:numberpredictionloss}
p(v_i \vert q, r,\mathcal{V}_r )= \frac{\exp \sigma( \textbf{W}_{\rm pretrain}^{\top} (\textbf{v}_i))}{\sum_{v_j \in \mathcal{V}_r} \exp \sigma( \textbf{W}_{\rm pretrain}^{\top}(\textbf{v}_j))},
}
\normalsize

\noindent where $\textbf{W}_{\rm pretrain}$ with  $\Theta$ are trained by a cross-entropy loss. 
\begin{figure}[t]
	\centering
	\includegraphics[width=0.44\textwidth]{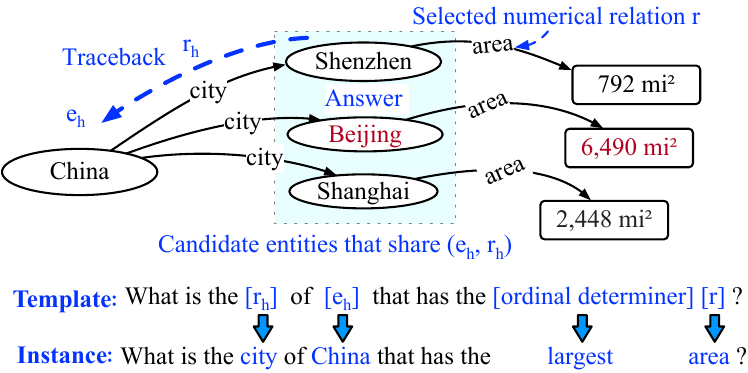}
	\caption{\label{fig:qapairs} A generated ordinal constrained QA pair.}
\end{figure}
We train NTL and NPL together.

\subsection{Data Augmentation}
\label{sec:QA_pairs}
\vpara{Issue of Insufficient Training.} The ordinal constrained QA pairs are usually much fewer than the normal QA pairs on the existing KBQA benchmarks, which may cause the insufficient training of the proposed model.
To verify our hypothesis, we conduct a quick experiment on the CWQ dataset. 
We reduce the ordinal constrained QA pairs by half in the original training dataset and re-train the model in Eq.\ref{eq:typeprediction}, which results in a drastic drop (-8\% Hits@1) of the ordinal constrained QA performance. 
To overcome this issue, we propose a template-based data augmentation method for increasing the ordinal constrained supervision signals and train the model on the union of the standard training data and the augmented data. 


\vpara{Generating Ordinal Constrained QA Pairs.} 
Our key idea is to automatically generate some pseudo ordinal constrained QA pairs by using a template-based method. We first randomly select a numerical relation $r$ from $\mathcal{G}$ and sample the entities $\{e\}$ and the corresponding numerical values $\{v\}$ from triplets in   $\mathcal{G}$ whose relation is $r$, \emph{i.e.}, each triplet of $r$, $e$, and $v$ satisfies $\{(e,r,v)\in \mathcal{G}\}$. Then starting from those entities $\{e\}$ as tails, we trace back to their first-order neighbors $\{(e_h,r_h)\}$ satisfying $\{(e_h,r_h,e)\in \mathcal{G}\}$. Finally, we choose the tail entities that share the same $(e_h,r_h)$ as the candidate entities $\mathcal{C}$. The textual question is constructed according to the template ``What is the [$r_h$] of [$e_h$] that has the [ordinal determiner] [$r$] ?''. The answer is selected from $\mathcal{C}$ with the numerical value that satisfies the instantiated ordinal determiner. Figure~\ref{fig:qapairs} shows a generated example, where ``area'' is the selected numerical relation $r$. Based on it, ``Shenzhen'', ``Beijing'', and ``Shanghai'' compose
the entities $\{e\}$ and the numerical values $\{v\}$ of their corresponding areas are collected. From these cities, we trace back to the common entity ``China'' and the relation ``city''. Finally, ``Beijing'' with the ``largest'' (the instantiated ordinal determiner) area ``6,490 mi$^2$'' is annotated as the answer.

\vpara{Overall Training Algorithm.} 
Algorithm~\ref{algo:training} presents the entire training process, where 
$\Phi$ of NSM (basic reasoning), $\Theta$ of \smodel (numerical reasoning), and $\Psi$ of the comprehensive reasoning module are parameters to be optimized. The overall algorithm contains two steps. The first step is pre-training $\Phi$ of basic reasoning and $\Theta$ of \model. The second step is training $\Psi$ of the comprehensive reasoning and fine-tuning $\Phi$ of basic reasoning with fixed $\Theta$ of \model. $\Theta$ of \smodel is only updated in pre-training and is frozen in numerical reasoning, which significantly improve the inference efficiency. More specifically, after pre-training $\Phi$ and $\Theta$, $\Phi$ excluding $\Theta$ is fine-tuned with $\Psi$, because the pre-trained $\Theta$ is good enough for capturing numerical properties, while fine-tuning $\Phi$ after being injected the number embeddings is for better answer selection. 


\section{Experiment}
\label{sec:experiment}
We design the experiments to answer the four questions: (1) Can the proposed \swmodel address ordinal constrained KBQA? (2) Is the framework adapted to other basic reasoning models? (3) Can pre-training boost the numerical skills? (4) Is the augmented ordinal constrained QA data helpful?

\subsection{Experimental Settings}
\vpara{Dataset and Evaluation Metrics.}
 We adopt two benchmarks, WebQuestionSP(WebQSP)~\cite{yih-etal-2016-value} and Complex WebQuestion 1.1(CWQ)~\cite{talmor-berant-2018-web} for evaluation. Since WebQSP contains inadequate ordinal constrained questions, we add LC-QuAD 2.0~\cite{dubey2017lc2} and KQA Pro~\cite{KQAPro}
 with only the ordinal constrained questions as additional datasets. Table~\ref{tb:dataset} shows the statistics. Hits@1 is used to evaluate whether the top-1 prediction is correct.


\begin{algorithm}[t]
\footnotesize{
      \caption{ \footnotesize Training Algorithm\label{algo:training}}
    \KwIn{KB, QIND, QGND, real and pseudo QA data.}
    \KwOut{Learned parameters $\Phi$, $\Theta$, and $\Psi$.}
     Initialize $\{\textbf{e}, \textbf{r}, \textbf{v}, \textbf{q}\}$ by frozen RoBERTa;\\
     
     \tcc{ \scriptsize{Pre-train the basic reasoning module}}
    Train $\Phi$ of NSM by Eq.~\ref{eq:basicpredict} on real QA data;\\
    \tcc{ \scriptsize{Pre-train \smodel on the question-irrelevant numerical dataset}}
    Train $\Theta$ of \smodel by Eq.~\ref{eq:numbertripletloss} and Eq.~\ref{eq:numberpredictionloss} on QIND;\\
    \tcc{ \scriptsize{Pre-train \smodel on the question-guided numerical dataset}}
    Train $\Theta$ \smodel by Eq.~\ref{eq:numbertripletloss} and Eq.~\ref{eq:numberpredictionloss} on QGND;\\
     \tcc{ \scriptsize{Subgraph, numerical relations/values retrieval}}
    Retrieve $(\mathcal{G}_q,\mathcal{R}_q,\mathcal{V}_q)$ for each $q$;\\
    \tcc{ \scriptsize{Basic Reasoning}}
    Use NSM to update $\{\textbf{e}\}$ by Eq.~\ref{eq:basicembedding};\\
    \tcc{ \scriptsize{Numerical Reasoning}}
    Use \smodel to update $\{\textbf{v}\}$ by Eq.~\ref{eq:transformerinput}-\ref{eq:attention};\\
    \tcc{ \scriptsize{Comprehensive Reasoning}}
    Fuse $\{\textbf{v}\}$ into the pruned $\{\textbf{e}\}$ by Eq.~\ref{eq:numberplugin}-\ref{eq:finalembedding};\\
    \tcc{ \scriptsize{End-to-end fine-tune NSM and train the comprehensive reasoning module}}
    Train $\Phi$ and $\Psi$ by Eq.~\ref{eq:typeprediction} on the union of the real and pseudo QA data.
}    
 \end{algorithm}

\begin{table}[t]
\newcolumntype{?}{!{\vrule width 1pt}}
\newcolumntype{C}{>{\centering\arraybackslash}p{2em}}
\caption{
	\label{tb:dataset} \#All/Ordinal QA pairs for train/val/test. (O) denotes that the dataset only contains ordinal constrained questions. 
}
\centering 
\renewcommand\arraystretch{1.0}
\small
\scalebox{0.95}{
\begin{tabular}{@{}l?ccc@{}}
\toprule
Dataset & Train & Validation & Test    \\ \midrule
WebQSP   & 2,848/58 & 250/4 & 1,639/39  \\
CWQ      & 27,639/1,435 & 3,519/189 & 3,531/197 \\
LC-QuAD 2.0 (O) & 213/213 & 48/48 & 110/110  \\
KQA Pro (O) & 759/759 & 305/305 & 462/462 \\\bottomrule
\end{tabular}
}
\end{table}


\vpara{Baseline Models.}
We compare with both our previous work~\cite{feng-etal-2021-pretraining-numerical} and several state-of-the-art baselines. The baselines include both the SP-based and the embedding-based methods, where the latter can be further divided into three categories given how the KB knowledge is utilized: memory table, subgraph, and whole KB.

\noindent \textbf{\textit{QGG}}~\cite{lan-jiang-2020-query}: is an SP-based method which parses a question into an executable query graph, where the multi-hop relation paths and the constraints are extended simultaneously.

\noindent \textbf{\textit{SPARQA}}~\cite{sun2020sparqa}: is also an SP-based method that invents a new skeleton grammar to guide the parsing results.

The following methods are all the embedding-based models.

\noindent \textbf{\textit{KV-Mem}}~\cite{miller-etal-2016-key}: transfers the KB triplets into key-value pairs and stores them into a memory table for KBQA.

\noindent \textbf{\textit{BAMnet}}~\cite{chen2019bidirectional}: is also built upon a memory network, but emphasizes the mutual interactions between questions and the underlying KB.

\noindent \textbf{\textit{EmbedKGQA}}~\cite{saxena-etal-2020-improving}: learns the question and entity embeddings on whole KB by maximizing the likelihood of all the (topic entity, question, answer) triplets.

\noindent \textbf{\textit{GRAFT-Net}}~\cite{sun-etal-2018-open}: is a subgraph-oriented method, which adopts a variant of GNN model to perform multi-hop reasoning on $\mathcal{G}_q$.

\noindent \textbf{\textit{NSM}}~\cite{He_2021}: is also a subgraph-oriented  method, which takes a generic neural state machine as the encoder and is currently the state-of-art embedding-based model. 

\noindent \textbf{\textit{\textbf{+Num}}}~\cite{feng-etal-2021-pretraining-numerical}: means injecting the previously proposed NumGNN and NumTransformer into an existing KBQA model.






\vpara{Training Details} are presented in Appendix~\ref{Experimental Settings}.

\begin{table}[t]
\newcolumntype{?}{!{\vrule width 1pt}}
	\newcolumntype{C}{>{\centering\arraybackslash}p{2em}}
	\caption{
		\label{tb:overall} QA performance on (all) and (ordinal) constrained test instances (Hits@1 by \%). NT- denotes the KBQA model is injected by \model. 
	}
	\centering 
	\small
	\renewcommand\arraystretch{1.0}
\scalebox{0.95}{
\begin{tabular}{@{}l@{ }?cc?cc@{}}
\toprule
\multirow{2}{*}{Model} & \multicolumn{2}{c?}{WebQSP} & \multicolumn{2}{c}{CWQ} \\  & All & Ordinal & All & Ordinal  \\\midrule
\multicolumn{5}{c}{SP-based Models} \\\midrule
SPARQA  & - & - & 31.6 & 6.3 \\
QGG  & \textbf{73.0} & \textbf{61.2} & 36.9 & 24.9 \\
\midrule
\multicolumn{5}{c}{Embedding-based Models} \\\midrule
KV-Mem &46.6 & 33.3 &18.4 & 11.7\\
BAMnet & 55.6 & 41.0 & - & - \\
EmbedKGQA   & 46.0  & 35.4    & 32.0 & 20.0    \\
GRAFT-Net   & 66.4  & 28.4    & 36.8  & 19.3      \\
GRAFT-Net + Num  & 67.4  & 43.2   & 37.3  & 25.8    \\
NT-GRAFT-Net ({\bf Ours}) & 67.2 & 41.0 & 37.3& 25.8  \\ 
NSM   & 68.5 & 33.3     & 46.3  & 24.4     \\
NSM + Num & 68.6 & 38.5 & 47.4 & 28.4 \\
\textbf{\wmodel}  ({\bf Ours}) & 69.1    & 46.2 & \textbf{48.9}& \textbf{42.1}      \\
 \midrule
\multicolumn{5}{c}{Ablation Studies}\\
\midrule
\swmodel w/o SAM   &68.2 & 38.5  & 47.8 & 35.5  \\
\swmodel w/o SNE   &68.7 & 43.6 &48.3 & 40.6 \\
\swmodel w/o QGND       &68.7  & 41.0 & 48.3  & 41.1      \\
\swmodel w/o QIND       & 68.1 & 38.5 & 47.7  & 36.0         \\
\swmodel w/o NPL       & 68.4 & 41.0 & 48.3   & 37.6       \\
\swmodel w/o NTL      & 68.8  & 43.6 & 48.4   & 41.6      \\
\swmodel w/o Pre-train  & 68.6 & 38.5 & 47.0  & 28.9      \\
\swmodel w/o DA  & 68.6  & 41.0  & 47.2   & 30.9       \\
\bottomrule
\end{tabular}
}
\end{table}

\subsection{Overall QA Performance}
Table~\ref{tb:overall} presents Hits@1 of all the methods. 
We observe that the performance of QGG and SPARQA on CWQ isn't as good as that on WebQSP, even is worse than the embedding-based models, because CWQ contains more complex questions, whose accurate logic forms are difficult to be parsed when the ground truth of the logic forms are missing or low-quality. 
On the contrary, if we inject the proposed \smodel into NSM---the embedding-based SOTA model, an obvious performance increase on both the datasets can be observed, \emph{i.e.}, NT-NSM improves 12.9\% Hits@1 on WebQSP (Ordinal) and 17.7\% on CWQ (Ordinal) compared with NSM. 
Table~\ref{tb:kqapro} demonstrates the same trend on LC-QuAD 2.0 and KQA Pro. Note that the original performance of GRAFT-Net and NSM on KQA Pro is unsatisfactory because this dataset contains particularly complex questions (Cf. Table~\ref{tb:ordinaldataset} in Appendix for some examples). Note the results of SPARQA on WebQSP and BAMNet on CWQ are not reported, as SPARQA needs to annotate special logic forms and BAMNet depends on subgraphs with entity types. 
We also observe that our newly proposed NT-NSM w/o DA (\emph{i.e.}, \swmodel removing data augmentation) performs better than our previously proposed NSM + Num on the same training dataset, which indicates that this newly proposed model is a more concise but effective choice.
The above results demonstrate that \swmodel can solve the task of ordinal constrained KBQA.

\begin{table}
\newcolumntype{?}{!{\vrule width 1pt}}
	\newcolumntype{C}{>{\centering\arraybackslash}p{2em}}
	\caption{
		\label{tb:kqapro} Hits@1 (\%) on LC-QuAD 2.0(O) and KQA Pro(O). 
	}
	\centering 
	\small
	\renewcommand\arraystretch{1.0}
\scalebox{0.8}{
\begin{tabular}{@{}c?cccc@{}}
\toprule
Model &GRAFT-Net&NT-GRAFT-Net&NSM & \textbf{\wmodel} \\ \midrule
LC-QuAD 2.0(O)  & 28.2 & 34.5 & 33.6  & \textbf{39.1} \\
KQA Pro(O) & 10.8 & 14.3 & 8.9  & \textbf{28.6} \\
\bottomrule
\end{tabular}}
\end{table}

\noindent \textbf{Effect of Self-attention Mask Matrix (SAM).} Table~\ref{tb:overall} indicates that by removing SAM from \model, 
Hits@1 of \swmodel significantly drops by 6.6-7.7\% on both the datasets. 
This is expected, as SAM helps distinguish numbers of different scales. 

\noindent \textbf{Effect of Special Number Embedding (SNE).}
Table~\ref{tb:overall} indicates that Hits@1 drops 1.5-2.6\% when changing SNE,  which concatenates the [S]'s and [E]'s embeddings, to the [CLS]'s embedding. The result indicates the effectiveness of SNE method. 

\begin{figure}[t]
	\centering
	\subfigure[Entity pruning]{\label{subfig:threhold}
		\includegraphics[width=0.23\textwidth]{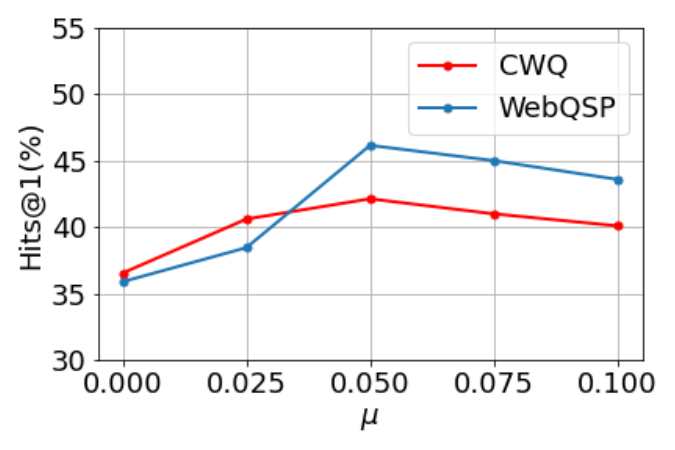}
	}
	\hspace{-0.1in}
    \subfigure[Augmented QA data]{\label{subfig:genrated}
		\includegraphics[width=0.23\textwidth]{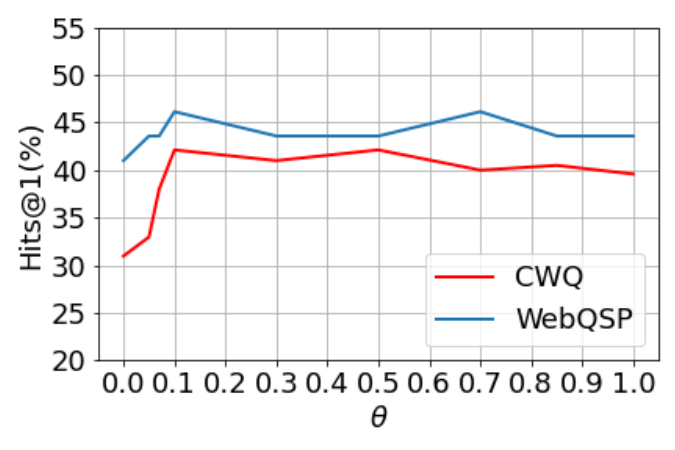}
	}
	\hspace{-0.1in}
	\subfigure[\#QIND]{\label{subfig:pretrainQGND}
		\includegraphics[width=0.22\textwidth]{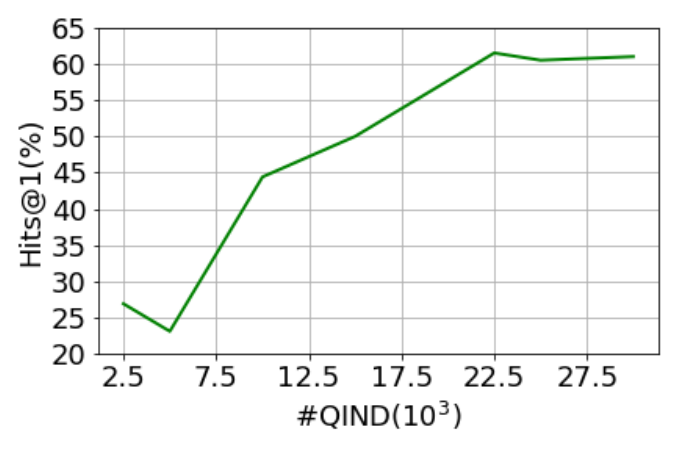}
	}
	\hspace{-0.09in}
	\subfigure[\#Numbers in QIND]{\label{subfig:number}
		\includegraphics[width=0.24\textwidth]{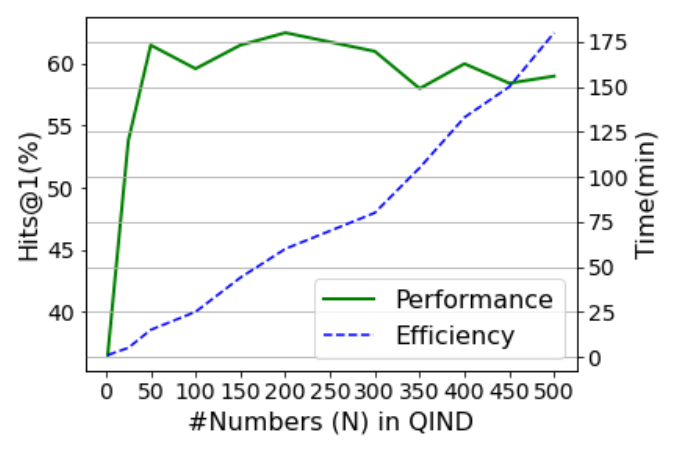}
	}
	
	\caption{\label{fig:pretrain} Effect of (a) $\mu$ (threshold for entity pruning); (b) $\theta$ (proportion of the augmented questions to the original data); (c) \#QIND (the size of QIND); (d) \#numbers in QIND (the number of numbers in an instance of QIND). (a) and (b) report the QA performance. (c) and (d) reports the direct pre-training performance. 
	}
\end{figure}

\noindent \textbf{Effect of Entity Pruning.} We vary the threshold $\mu$ that controls the entities being injected with the number embeddings and show different QA performances in Figure~\ref{subfig:threhold}. The best performance is achieved when $\mu$ equals 0.05 on both the datasets, because too small $\mu$ may prune the answer and too large $\mu$ may include much noise.

\vpara{Adaptation to Other Basic Reasoning Models.}
We replace NSM with GRAFT-Net in \swmodel to create NT-GRAFT-Net and obtain 6.5-12.6\% Hits@1 improvement on GRAFT-Net (Table~\ref{tb:overall}). However, only about 2\% is improved when injecting \smodel into KV-Mem and EmbedKGQA, possibly because: (1) they reason on the whole KB instead of a subgraph, which increases the difficulty of pruning noisy entities; (2) their parameters are inadequate for fine-tuning entity embeddings into the same space with number embeddings. On the contrary, NSM, and GRAFT-Net both reason on subgraphs by rich parameterized GNN encoders.

\subsection{Pre-training and Augmentation}
\vpara{Effect of Pre-training.}
We consider the following variants, aiming to investigate the contribution of the two generated datasets for pre-training and two pre-training loss functions. 
Specifically, we remove the whole pre-training process of \smodel (w/o pre-train) and also remove each component (w/o QGND, w/o QIND, w/o NPL, and w/o NTL). The results in Table~\ref{tb:overall} show that: (1) the pre-training tasks allow \swmodel to better acquire the numerical skills. (2) Both QIND and QGND present positive guidance for \wmodel. 
QIND contributes more than QGND, because QIND is fully auto-generated, which can be much larger than QGND that is restricted by the size of the real ordinal constrained QA pairs. (3) Both NTL and NPL present positive guidance for \wmodel. NPL contributes more than NTL, because NPL is more matched with the downstream QA task to find the answer satisfying the ordinal constraint. 

\noindent \textbf{Effect of \#QIND.} We adjust the size of QIND from 2,500 to 30,000 with an interval of 2,500 and present the direct pre-training performance on QIND in Figure~\ref{subfig:pretrainQGND}. Specifically, we evaluate the ability to predict the right number that satisfies the ordinal constraint of a question, which is designed in the same way as NPL, using QGND's test set. 
The performance becomes stable when \#QIND is larger than 22,500, which is selected as the QIND size for pre-training.

\noindent \textbf{Effect of \#Numbers.} We adjust $N$, the number of numbers in a QIND instance from 2 to 500 with an interval of 50 and present the direct pre-training performance with the corresponding training time of an epoch in Figure~\ref{subfig:number}. To consider efficiency and performance, 50 is adequate for pre-training.

\noindent \textbf{Effect of Data Augmentation.}  Table~\ref{tb:overall} indicates that by removing data augmentation (DA), Hits@1 of \swmodel significantly drops by 5.2-11.2\% on both datasets. We vary the proportion of the augmented ordinal constrained QA pairs to the original training set from 0\% to 100\% with an interval 10\%, and present the corresponding QA performance in Figure~\ref{subfig:genrated}. 
The performance gets stable when the generated data is larger than 10\% of the original data. We conjecture that the template for generating the dataset is somewhat specialized, which is unfavorable to the model generalization. More various templates could be added in the future.

\section{Conclusion}
This paper proposes an embedding-based KBQA framework that considers numerical reasoning for addressing ordinal constrained KBQA. The core is a \underline{N}umerical\underline{T}ransformer (NT) for learning number embeddings and a comprehensive reasoning module for injecting the number embeddings into the basic entity embeddings learned by any embedding-based KBQA model. To enable better training, we propose two numerical-oriented pre-training tasks as well as a template-based data augmentation strategy. Extensive experiments demonstrate via the training strategy, the proposed \swmodel can achieve substantial improvements on numerical reasoning compared with the baselines.


\bibliographystyle{acl_natbib}

\clearpage
\appendix
\section{Experimental Details}

\subsection{The Ordinal Constrained Question Examples}
\label{Sample Ordinal Constrained Questions}
Table~\ref{tb:ordinaldataset} demonstrates several ordinal constrained question examples for the four KBQA benchmarks, which indicates WebQSP is the easiest dataset and KQA Pro is the hardest one.

\subsection{Experimental Settings}
\label{Experimental Settings}

\vpara{General Setting.}
We run experiments with one RTX 3090(24G) GPU on the server with 256G physical memory and 8T disk size. We adopt RoBERTa-base~\cite{liu2019roberta} as the basic PLM model, which consists of 12 layers, 768-d hidden size, 12 attention heads, and 110M parameters in total. Due to the scarce ordinal labels on the validation set of WebQSP , model selection is performed on WebQSP’s training set instead. The learning rate is 0.0001. 

\vpara{Numerical Reasoning.}
For each question, we retrieve top-$K$ ($K$=3) relevant numerical relations. 
We unify the units of the numbers belonging to the same numerical relation. In \model, the layer size is 2 and the head size is 8.
 We restrict the maximal $N$, the number of the input numbers of \smodel within 50 numbers for WebQSP/CWQ dataset and 500 for KQA Pro/LC-QuAD 2.0 dataset according to their original data sizes. 

\vpara{Basic Reasoning.}
All basic reasoning models adopt the same settings as the original papers. For the GRAFT-Net and NSM, the  subgraph is retrieved following GRAFT-Net's method to extract the neighborhood relation triplets within two hops of the topic entity and then perform the personalized PageRank~\cite{10.1145/511446.511513} to keep the most relevant entities to question.

\vpara{Comprehensive Reasoning.}
We train the question type classifier on both the original CWQ training set and the augmented ordinal constrained questions, leading to almost 100\% accuracy. The threshold $\mu$ for entity pruning is set as 0.05.

\vpara{Pre-training.}
For pre-training, we generate 40,000 and 645 instances for QIND and QGND respectively, which are partitioned into 24,000/8,000/8,000 and 500/60/85 for training/validating/testing. Questions in QGND do not appear in any test dataset of the four KBQA benchmarks. The margin $\epsilon$ is set as 0.5.  
We restrict $N$, the number of the input numbers of \smodel within 2 to 50 numbers in pre-training. For each training instance, we determine the concrete $N$ by sampling a number from the above scope [2,50].

\vpara{Data Augmentation.}
We generate 2,700 pseudo ordinal constrained instances for CWQ and 280 for WebQSP, about 10\% of the original training data.

\vpara{Training Time.}
For pre-training, an epoch with batchsize 300 takes 60 seconds and it converges within 15 epochs for the first pre-training task (QIND). An epoch with batchsize 300 takes 45 seconds and it converges within 15 epochs for the second pre-training task (QGND). For training, an epoch with batchsize 40 takes 60/600/20/1 seconds on WebQSP/CWQ/KQA Pro/LC-QuAD 2.0 and the model converges within 50 epochs for NSM. An epoch with batchsize 40 takes 65/850/660/25/2 seconds on WebQSP/CWQ(with data augumentation)/CWQ/KQA Pro/LC-QuAD 2.0 and the model converges within 50 epochs for NT-NSM.

\subsection{Direct Pre-training Performance}
\label{Direct pre-training performance}
The results in Table~\ref{tb:directpretraining} demonstrate the effectiveness of different components by the direct pre-training performance, where QIND, NPL and SAM show the most critical effect.

\begin{table}
\newcolumntype{?}{!{\vrule width 1pt}}
	\newcolumntype{C}{>{\centering\arraybackslash}p{2em}}
	\caption{
		\label{tb:directpretraining} Direct pre-training performance of \smodel (NT) (\%).-XX means removing the component.}
	\centering 
	\small
	\renewcommand\arraystretch{1.0}
\scalebox{0.825}{
\begin{tabular}{@{}c?ccccccc@{}}
\toprule
Model & NT & -QIND  &  -QGND &   -NPL &   -NTL & -SAM & -SNE\\ \midrule
Hits@1 & 88.5 & 44.3 & 61.5  &36.5 & 84.6 & 44.2 &86.5 \\
\bottomrule
\end{tabular}}
\end{table}








\begin{table*}[t]
\newcolumntype{?}{!{\vrule width 1pt}}
\newcolumntype{C}{>{\centering\arraybackslash}p{35em}}
\caption{
	\label{tb:ordinaldataset} The Sampled Ordinal Constrained Questions.
}
\centering 
\renewcommand\arraystretch{1.0}
\small
\scalebox{0.95}{
\begin{tabular}{@{}l?C@{}}
\toprule
Dataset & Examples \\ \midrule
\multirow {2}{*}{WebQSP}   &  What was pink floyd 's first album?  \\
   &  Which year did lakers win their first championship?  \\\midrule
\multirow {2}{*}{CWQ}      & Which TV program does Nick Cannon play in, that has the largest amount of episodes? \\ 
&What is the oldest sports team that Hank Baskett played for in Nov 2010? \\ \midrule
\multirow {2}{*}{LC-QuAD 2.0 (Ordinal)}      & Which BMW M20 has the highest straight six torque engine? \\ 
&What battery power station has the highest amount of energy storage capacity? \\ \midrule
\multirow {3}{*}{KQA Pro (Ordinal)}  & Among the independent cities that share a border with York County (the one that shares border with Williamsburg), which one has the largest area?  \\
&Which is the shortest among the animated feature films whose genre is buddy film? \\ \bottomrule
\end{tabular}
}
\end{table*}
\end{document}